%% file: main.tex
\documentclass[sigconf, author]{acmart}

\usepackage{lib}

\usepackage[utf8]{inputenc} 
\usepackage[T1]{fontenc}    
\usepackage{hyperref}       
\usepackage{url}            
\usepackage{booktabs}       
\usepackage{amsfonts}       
\usepackage{nicefrac}       
\usepackage{microtype}      
\usepackage{xcolor}         

\usepackage[ruled]{algorithm2e}
\usepackage{amsmath}
\usepackage{colortbl}
\usepackage{enumitem}
\usepackage{float}
\usepackage{graphicx}
\usepackage{hypcap}
\usepackage{listings}
\usepackage{mathtext}
\usepackage{pgfplots}  
\usepackage{rotating,tabularx}
\usepackage{siunitx}
\usepackage{wrapfig}  
\usepackage{bm} 

\AtBeginDocument{%
  }

\setcopyright{acmlicensed}

\acmConference[KDD CJ’24, August, 2024, Barcelona]{3rd Workshop on End-to-End Customer
Journey Optimization.}{}

\acmDOI{}

\begin{document}

\title{Efficient Feature Interactions with Transformers: Improving User Spending Propensity Predictions in Gaming}

\author{Ved Prakash}
\email{ved.prakash@dream11.com}
\affiliation{%
  \institution{Dream11}
  \country{India}
}
\authornote{All authors contributed equally to this research.}

\author{Kartavya Kothari}
\email{kartavya.kothari@dream11.com}
\affiliation{%
  \institution{Dream11}
  \country{India}
}

\renewcommand{\shortauthors}{Ved et al.}


\begin{abstract}
  \input{text/abstract}
\end{abstract}




\maketitle

\section{Introduction}\label{sec:introduction}

\input{text/introduction}


\section{Related Work}\label{sec:related-work}
\input{text/related-work}

\section{Problem-formulation}\label{sec:definition}
\input{text/problem-formulation.tex}

\section{Method}\label{sec:method}
\input{text/method}

\section{Experiments and Results}\label{sec:experiments-and-results}
\input{text/experiments}

\section{Discussion}\label{sec:discussion}
\input{text/discussion}

\section{Conclusion}\label{sec:conclusion}
\input{text/conclusion}



\bibliographystyle{ACM-Reference-Format}
\bibliography{references}

\appendix







\end{document}

%% file: text/abstract.tex

Dream11 is a fantasy sports platform that allows users to create their own virtual teams for real-life sports events.
We host multiple sports and matches for our 200M+ user base. In this RMG (real money gaming) setting, users pay
an entry amount to participate in various contest products that we provide to users. In our current work, we discuss the problem of 
predicting the user's propensity to spend in a gaming round, so it
can be utilized for various downstream applications. e.g. Upselling users by incentivizing them marginally as per
their spending propensity, or personalizing the product listing based on the user's propensity to spend.

We aim to model the spending propensity of each user based on past transaction data. In
this paper, we benchmark tree-based and deep-learning models that show good results on structured data, and we propose a new architecture change that
is specifically designed to capture the rich interactions among the input features. We show that our
proposed architecture outperforms the existing models on the task of predicting the user's propensity to spend in a
gaming round. Our new transformer model surpasses the state-of-the-art FT-Transformer, improving MAE by 2.5\% and MSE by 21.8\%.

%% file: text/introduction.tex
Fantasy sports have revolutionized the gaming industry, blending real-world sports events with virtual team creation, allowing users to engage in competitive and immersive experiences. Dream11 offers users the opportunity to participate in various contest formats by paying an entry fee, with the potential to win based on the performance of their chosen virtual teams. Research indicates that users' spending behavior in fantasy sports is influenced by various factors such as engagement level, 
perceived skill, and competition thrill \citet{suh2010examining} \citet{doi:10.1080/14459795.2014.881904}. In this dynamic landscape, predicting a user's propensity to spend in a gaming round is paramount for enhancing user engagement. Accurately predicting a user's propensity to spend in a gaming round is critical for optimizing user engagement through personalized product recommendations and targeted incentives, thereby enhancing overall user experience. The current work focuses on modeling user spending propensity through the analysis of extensive transaction data, encompassing gaming participation, spending, winnings, demographic details, and other pertinent features.

Given the large tabular datasets with millions of rows and hundreds of features, the choice of modeling technique becomes crucial. Gradient Boosting Decision Trees (GBDT) have traditionally been the go-to models for handling such tabular data effectively \citet{gorishniy2023revisiting}. However, the increasing volume and complexity of tabular data have challenged the efficacy of traditional machine-learning models. While these models can handle large datasets up to a limit, they become computationally expensive to train as the data size and feature space grow. Consequently, traditional ML models may not fully leverage the vast amount of available data, and intricate feature relationships

In contrast, deep learning models have demonstrated promise in capturing complex patterns, particularly in fields such as computer vision, natural language processing, and speech recognition given large-scale data. Despite their dominance in these areas, deep learning based methods are still not common on tabular data \citet{shwartz2022tabular}. One reason for this is the inherent challenges posed by heterogeneous tabular data, which often includes a mix of continuous, categorical, and ordinal features. Additionally, GBDT models have long been favored for tabular data due to their ease of use and reliable performance.

For instance, \citet{borisov2022deep} found that deep learning models can be more efficient in terms of inference time, particularly for large-scale data, compared to GBDT models. \citet{hestness2017deep} also showed that the performance of deep learning models improves with the increase in dataset size.

Despite these advantages, deep learning models face several challenges in tabular data modeling. The non-rotationally-invariant nature of tabular data features and the presence of uninformative data can hinder network generalization \citet{chen2023excelformer}. Moreover, neural networks are often less robust to uninformative features, which are common in tabular datasets \citet{grinsztajn2022tree}.

To address these challenges, we propose a novel deep-learning architecture specifically designed for capturing richer feature interactions present in tabular datasets. Our main contributions are as follows:

\begin{enumerate}
  \item We propose a deep learning architecture that improves generalization by employing joint training to capture proximity-aware contextual relationships between input features.
  \item We establish a fair comparison between deep learning models and GBDT models based on their architectures and
        performance on real-world data.
\end{enumerate}


%% file: text/related-work.tex
Deep learning is scalable and offers advantages in terms of maintenance, MLOps, reusability and time to production. 
Recent work on attention-based architectures being adapted for tabular data is promising. One such work is tabtransformer \citet{tabtransformer} which focuses on 
using multi-head self-attention mechanism to model the relationships between categorical features, transforming them into contextual embeddings.
 The transformed categorical features are then concatenated with the continuous features and passed through a feed-forward neural network to 
 make predictions. This work was extended by \citet{gorishniy2022embeddings}. They evaluate 
 the performance of several deep learning models for tabular data and proposed FTTransformer, a transformer-based architecture for tabular data. 
 FT Transformer extended the embeddings for numerical features as well to improve upon the Tabtransformer. 
\citet{somepalli2021saint} propose using the intersample attention mechanism 
to model relationships between samples of a batch. This intersample attention is effective when there are few training data points
 coupled with many features(common in biological datasets).  

Building on this work, we propose a novel deep-learning architecture for tabular data that can improve generalization by employing
 joint training to capture proximity-aware contextual relationships between input features. 
We evaluate the deep learning models on real-world data in the fantasy sports domain.

\citet{mcelfresh2024neural} conducted an exhaustive study and uncovered insights into the performance of 
deep learning models for tabular data. They have benchmarked 176 datasets, with only one dataset comprising approximately 
1 million rows and 10 features. In our domain, datasets typically contain millions of rows and hundreds of features. 
Consequently, we investigate the performance of deep learning models on such large-scale datasets.




%% file: text/problem-formulation.tex







Our objective is to predict a user's propensity to spend in a gaming round, aiding in upselling and personalizing product listings. We model spending propensity using past transaction data, including transaction history, gaming rounds participated in, amounts spent, winnings, demographic data, and other features.

Let \( P \) represent the user's spending propensity. We aim to predict \( P \) using the following features:

\begin{itemize}
    \item \( U \): User features, including demographics \( U_d \), aggregated past winnings \( U_w \), wallet balance \( U_b \), etc.
    \item \( R \): Game features, including type of game \( R_g \), expected active round users \( R_u \), etc.
    \item \( C \): Proximity-aware features - Game features of rounds before and after the current round, including features \( C_t \) from \( t-5 \) to \( t+5 \) where \( t \) is the current round
\end{itemize}

The data can be represented as tuples \((U, R, C, P)\). Our goal is to learn a function \( f \) such that:

\[ P = f(U, R, C) \]

where

\[ U = \{U_d, U_w, U_b, \ldots\} \]
\[ R = \{R_g, R_f, R_u, \ldots\} \]
\[ C = \{R_{t-5}, \ldots, R_{t+5}, \ldots\} \]

Given a large dataset \( D \) of these tuples, 

\[ D = \{(U_i, R_i, C_i, P_i)\}_{i=1}^{N} \]

where \( N \) is the number of data points, we aim to accurately learn the function \( f \) for predicting spend propensity.

\[
\hat{P}_i = f(U_i, R_i, C_i) \quad \forall i \in \{1, 2, \ldots, N\}
\]

The learned function \( f \) should minimize the prediction error, typically measured by a loss function \( \hat{L} \):

\[
    \hat{L} = \min_f \sum_{i=1}^{N} L(P_i, \hat{P}_i)
\]

where \( \hat{P}_i \) is the predicted spending propensity and \( P_i \) is the actual spending propensity.

%% file: text/method.tex


\subsection{Proximity-Aware Contextual Transformer}


In this section, We introduce Proximity-Aware Contextual Transformer - an adaptation of FT Transformer. Our adaptation
is motivated by the fact that the existing deep learning architectures are not effective in the presence of a large
number of features \citet{borisov2022deep}. We provide structure to the neural networks for learning from feature
relationships. In a nutshell, our model transforms additional feature relationships between input features into Contextual
Proximity-Aware embeddings and applies a stack of transformer models over embeddings.

The "Proximity-Aware Contextual Transformer relationship" in our model refers to statistical correlations(e.g., correlation between user age and spending),
 temporal patterns(e.g., user spending patterns over different times of the day), and contextual proximity(e.g.,
 clusters of similar spending behaviors) in feature data. By embedding these relationships,
the model can more effectively learn and predict outcomes based on the structured interplay of features.


\textbf{Notation.}
In this work, we consider supervised learning problems.
Let $(\bm{x}, y)$ denote a feature-target pair, where \[\bm{x} \equiv \{\bm{x}_{\text{cat}}, \bm{x}_{\text{cont}}, \bm{x}_{\text{context}}\}\] 
$\bm{x}_{\text{cat}}$ denotes all the categorical features, $\bm{x}_{\text{cont}} \in \mathbb{R}^{c}$ denotes $c$ continuous features 
and $\bm{x}_{\text{context}} \in \mathbb{R}^{s}$ denotes $s$ proximity-Aware contextual features. Let $\bm{x}_{\text{cat}} \equiv \{x_1, x_2, \cdots, x_m\}$ with
 each $x_i$ being a categorical feature, for $i \in \{1,\cdots, m\}$.
The dataset is split into three disjoint subsets: $D = D_{train} \ \cup \ D_{val}\ \cup \ D_{test}$, where $D_{train}$ is used for training, $D_{val}$ is 
used for early stopping and hyperparameter tuning, and $D_{test}$ is used for the final evaluation.



\captionsetup{
  font=normalsize, 
  labelfont=normalfont, 
  textfont=normalfont 
}

\begin{figure*}[ht]
  \centering
  \includegraphics[width=0.8\linewidth]{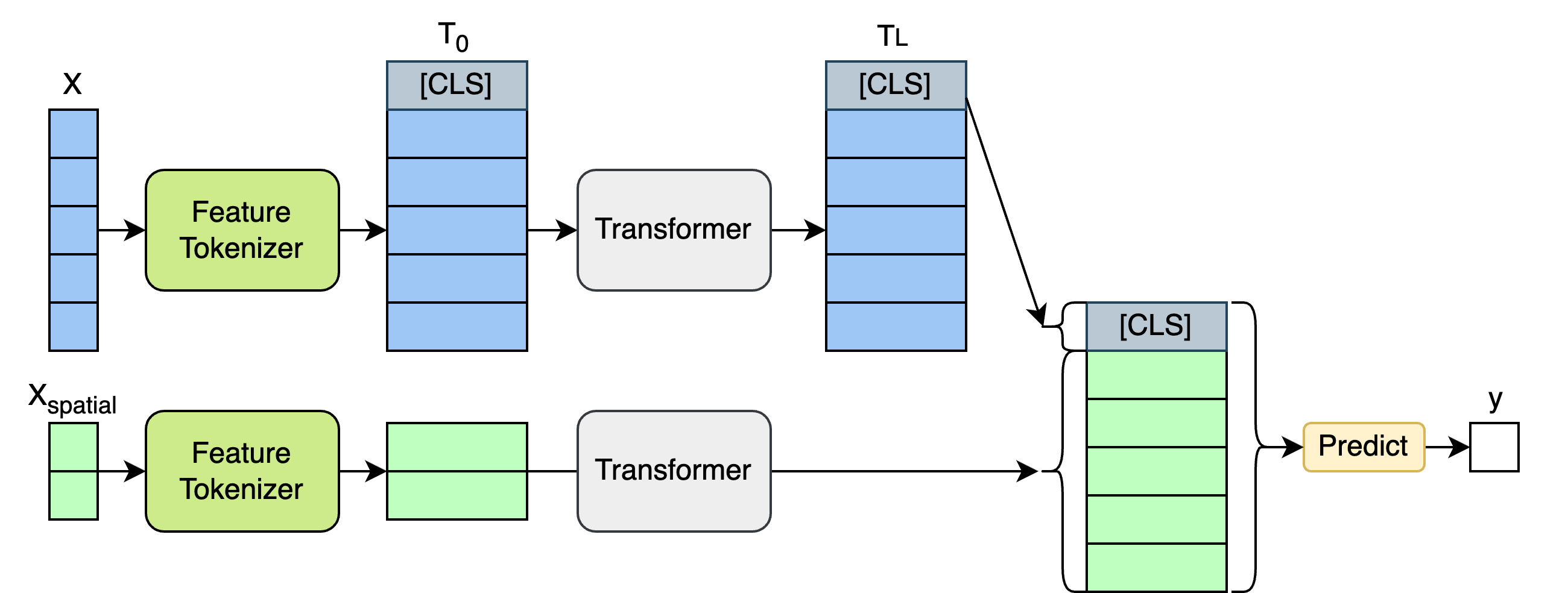} 
  \caption{The \architecture\ architecture. Firstly, \tokenizer\ transforms features to embeddings. The embeddings are then processed by the Transformer module and the final representation of the \CLS\ token is used for prediction.}
  \label{fig:arch}
\end{figure*}

\begin{figure}[ht]
  \centering
  \includegraphics[width=0.75\linewidth]{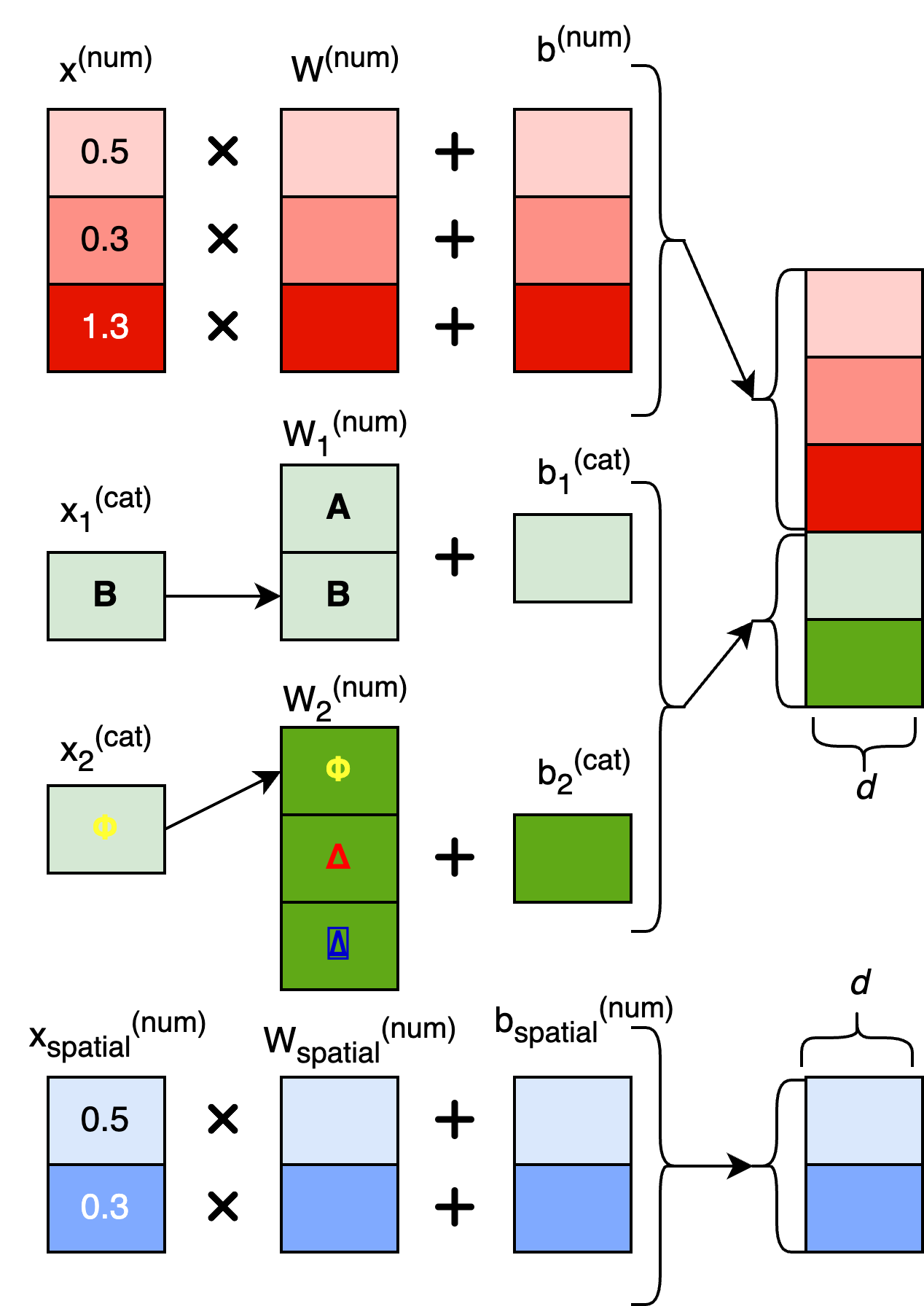}
  \caption{(a) \tokenizer; in the example, there are three numerical and two categorical features; (b) One Transformer layer.}
  \label{fig:blocks}
\end{figure}

\textbf{Feature Tokenizer.} \tokenizer\ module transforms input features $x$ to embeddings $T \in \R^{k \times d}$. The embedding for a given feature $x_j$ is computed as follows:

$$
  T_j = b_j + f_j(x_j) \in \R^d \qquad f_j: \X_j \rightarrow \R^d.
$$
where $b_j$ is the $j$-th \textit{feature bias}, $f^{(num)}_j$ is implemented as the element-wise multiplication with the vector \mbox{$W^{(num)}_j \in \R^d$} and $f^{(cat)}_j$ is implemented as the lookup table \mbox{$W^{(cat)}_j \in \R^{S_{j} \times d}$} for categorical features. Overall:
\begin{alignat*}{2}
   & T^{(num)}_j  = b^{(num)}_j + x^{(num)}_j \cdot W^{(num)}_j                                                                               &  & \in \R^d,            \\
   & T^{(cat)}_j  = b^{(cat)}_j  + x^{(cat)}_j\cdot  W^{(cat)}_j                                                                              &  & \in \R^d,            \\
   & T            = \mathtt{stack} \left[ T^{(num)}_1,\ \ldots,\ T^{(num)}_{k^{(num)}},\ T^{(cat)}_1,\ \ldots,\ T^{(cat)}_{k^{(cat)}} \right] &  & \in \R^{k \times d}.
\end{alignat*}





\subsection{Benchmark models}
\label{sec:other-models}
We benchmark our proposed approach against existing models designed specifically for tabular data.

\begin{itemize}[nosep]

  \item \textbf{MLP} \citet{gorishniy2023revisiting}. A multi-layer perceptron.
  \item \textbf{ResNet} \citet{gorishniy2023revisiting}. A deep residual network.
  \item \textbf{Tabtransformer} \citet{tabtransformer}. A transformer-based architecture for tabular data.
  \item \textbf{FT Transformer} \citet{gorishniy2023revisiting}. The SOTA model for self attention based tabular data modeling.
  \item \textbf{XGBoost} \citet{xgboost}. One of the most popular GBDT implementations.
  \item \textbf{CatBoost} \citet{catboost}. GBDT implementation that uses oblivious decision trees \citet{odt} as weak learners.
\end{itemize}

%% file: text/experiments.tex
In this section, we compare DL models to each other as well as to GBDT.


\subsection{Scope of the comparison}

In our work, we concentrate on the relative performance of various architectures and do not utilize model-agnostic deep learning practices such as pretraining, additional loss functions, data augmentation, distillation, learning rate warmup, learning rate decay, among others. Although these practices may enhance performance, our evaluation focuses on the impact of the inductive biases introduced by different model architectures.






For each dataset, we have used the same train-test split and the same evaluation metric for different architectures. 
You can find the user round spends data statistics in Table \ref{table:data_overview}.

\begin{table}[h!]
    \centering
    \begin{tabular}{lrrr}
    \toprule
     & \textbf{Train} & \textbf{Val} & \textbf{Test} \\
    \toprule
    Tuples & 7,074,749 & 1,165,011 & 831,583 \\
    Categorical Features & \multicolumn{3}{c}{21} \\
    Proximity-aware contextual Features & \multicolumn{3}{c}{60} \\
    Numerical Features & \multicolumn{3}{c}{120} \\
    Task & \multicolumn{3}{c}{Regression} \\
    \bottomrule
    \end{tabular}
    \caption{Overview of User Round Spends Data}
    \label{table:data_overview}
    \end{table}

\subsection{Implementation Details}
\label{sec:Implementation Details}

\textbf{Data Preprocessing}
We apply standardization and normalization to continuous features. We do not apply any standardization or normalization to the target variable. 

\textbf{Hyperparameter search}
For hyperparameter search we employed Optuna library \cite{optuna} to run Bayesian optimization \cite{hp-tuning}. The optimal parameters selected based on performance on validation set, the test set is never used for hyperparameter search. We set budget for optimization in terms of iterations and time. 

\textbf{Evaluation}
For each tuned configuration, we run 10 experiments with different random seeds and report average performance.

\textbf{Categorical Features}
We apply label encoding for categorical features for Xgboost, lightgbm and catboost. For deep learning models, we use embeddings for categorical features. Embedding size is consistent across all models.

\subsection{Comparison with Other models}
\label{sec:Comparison with Other models}

To compare performance of different algorithms, we conduct experiments on tabular datasets. We evaluate Mean Absolute Error (MAE) and Mean Squared Error(MSE) metric for each algorithm. Results are summarized in Table \ref{tab:comparison}.
Based on both MAE and MSE, Proximity-Aware Contextual Transformer is the top-performing model, indicating it provides the most accurate 
predictions with the least error. 
While Tabtransformer and FT Transformer models demonstrate strong performance, Proximity-Aware Contextual Transformer surpasses them in both accuracy and consistency.
GBDT Baseline model is the least accurate according to both MAE and MSE metrics.

We choose two different evaluation metrics to understand model performance, a model with moderate but consistent errors will have a lower MAE but a higher MSE if another model has similar errors on average but fewer large errors.

Table \ref{tab:comparison} provides additional insights into the performance of the MLP model. While the model generally performs well for most data points, it exhibits a few outliers with significantly higher errors. These large errors can have a greater impact on the MSE metric compared to the MAE metric.

\begin{table}[htbp]
    \centering
    \caption{Comparison of Algorithms based on MAE and MSE}
    \label{tab:comparison}
    \begin{tabular}{|c|c|c|}
        \hline
        \textbf{Algorithm} & \textbf{MAE} & \textbf{MSE} \\
        \hline
        GBDT Baseline & $42.4$ & $500.35$ \\
        MLP  & $40.4$ & $362.53$ \\
        Resnet & $41.34$ & $397.48$ \\
        Tabtransformer & $37.99$ & $442.74$ \\
        FT Transformer & $38.07$ & $448.90$ \\
        \textbf{Proximity-Aware Contextual Transformer} & \textbf{37.13} & \textbf{351.01} \\
        \hline
    \end{tabular}
\end{table}

In Table \ref{tab:comparison}, we present the MAE and MSE values for each algorithm.



\subsection{Performance Analysis}

\subsubsection{How does MLP model perform ? Is it really a strong baseline ?} 
\citet{gorishniy2023revisiting} highlighted in their study that the basic MLP model can be a strong baseline.



The discrepancy between the MLP model’s MAE and MSE provides valuable insights:

\textbf{Error Distribution}: The relatively high MAE but low MSE indicates that while the MLP errors are larger on average (as reflected in the MAE), it has fewer extreme errors. The squared errors in MSE metric heavily penalizes larger errors, so a lower MSE suggests that the MLP model avoids making significant prediction mistakes, which is advantageous in applications where large errors are particularly costly.

\textbf{Model Consistency}: The MLP’s performance demonstrates consistency in prediction accuracy. Despite higher average error (MAE), the lower MSE indicates that the errors are more uniformly distributed without extreme outliers. This is an important characteristic, suggesting that the MLP model provides reliable performance across most predictions.

The MLP model MAE indicates that its average error is somewhat higher compared to the best-performing models, its low MSE highlights its strength in minimizing significant errors. This makes the MLP model a valuable option in scenarios where avoiding large prediction errors is crucial.

Overall, the MLP model performs well compared to the GBDT Baseline and Resnet models but is outperformed by the Tabtransformer, FT Transformer, and Proximity-Aware Contextual Transformer. However, the MLP’s ability to limit large errors, as evidenced by its low MSE, makes it a robust and reliable choice for many predictive tasks. This analysis underscores the importance of using both MAE and MSE metrics to gain a 
comprehensive understanding of a model’s performance, ensuring that both average accuracy and error distribution are considered.



\subsubsection{Does performance improve with dataset size?}  

\begin{table}[h!]
    \centering
    \begin{tabular}{|c|c|}
    \hline
    \textbf{Train Size \%} & \textbf{MAE} \\
    \hline
    0.05 & 38.56 \\
    \hline
    0.15 & 37.56 \\
    \hline
    0.3 & 37.54 \\
    \hline
    0.4 & 37.43 \\
    \hline
    0.5 & 37.13 \\
    \hline
    0.7 & 36.95 \\
    \hline
    0.8 & 36.86 \\
    \hline
    0.9 & 36.78 \\
    \hline
    \end{tabular}
    \caption{Performance of Proximity-Aware Contextual Transformer with increasing train sizes}
    \label{table:spatial_ft_transformer}
    \end{table}

\begin{figure}[ht]
    \centering
    \includegraphics[width=0.9\linewidth]{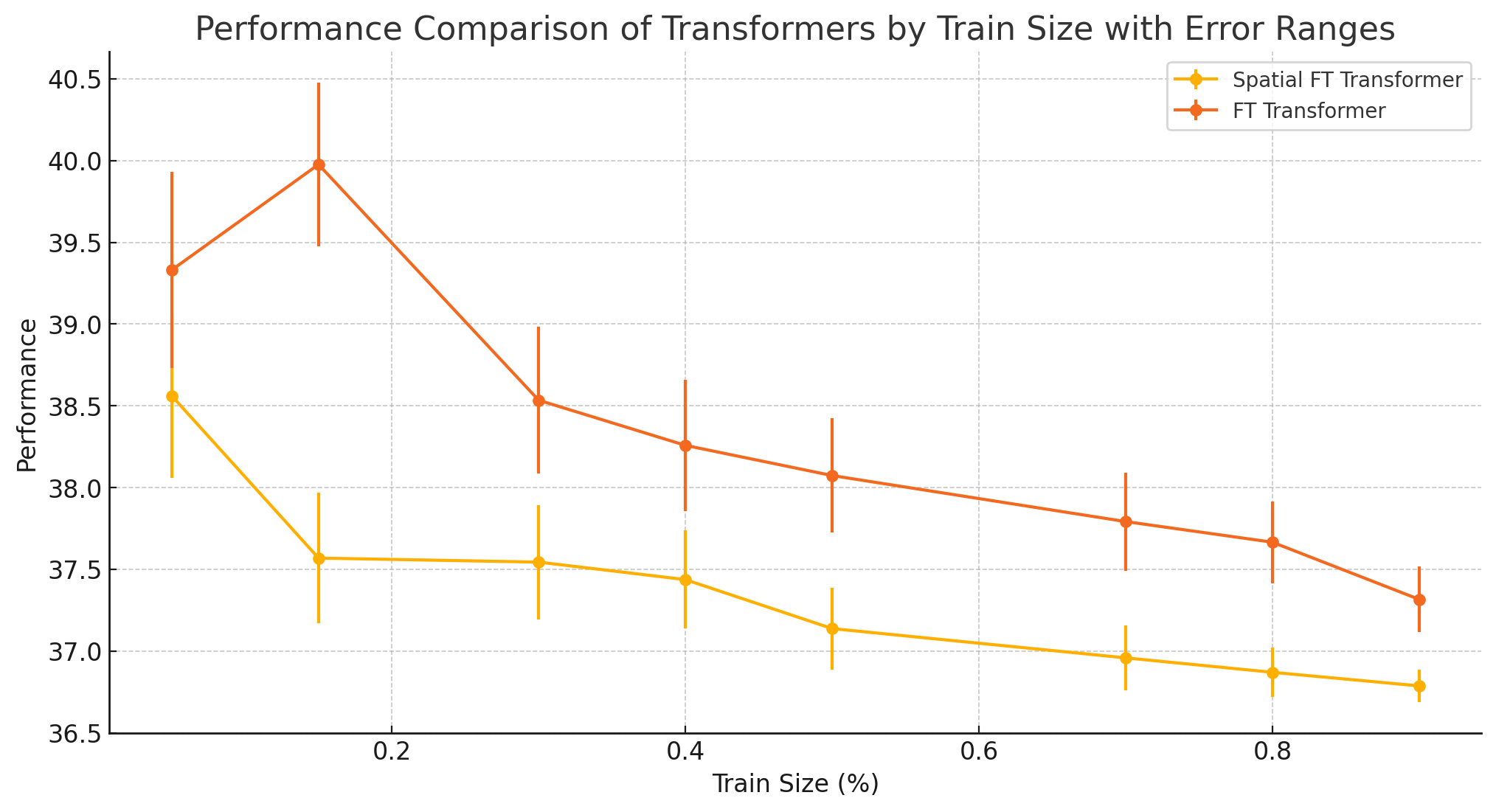}
    \caption{The graph compares the performance of the Proximity-Aware Contextual Transformer and the FT Transformer across different train
     sizes, represented as a fraction of n-observations. Error bars represent standard deviation from the mean across different runs, illustrating the variability due to the use of various random seeds.}
    \label{fig:maeperf}
\end{figure}

Performance does improve with the increase in the dataset size. We have experimented with different train sizes and 
observed that the performance of the Proximity-Aware Contextual Transformer improves with the increase in the dataset size. 
The results are summarized in Figure \ref{fig:maeperf}    

Proximity-Aware Contextual Transformer consistently outperforms the FT Transformer, particularly at smaller train sizes. This shows that model is able to converge faster than FT Transformer and attain the best performance of FT Transformer with smaller data size.

%% file: text/discussion.tex

In context of predicting user spending propensity on Dream11 platform, Proximity-Aware Contextual Transformer architecture demonstrates superior performance compared to both traditional GBDT models and other deep learning architectures.

\subsection{Interpretation of Experimental Results}
Experiments revealed that Proximity-Aware Contextual Transformer consistently outperformed other models in terms of Mean Absolute Error (MAE). This suggests that incorporating proximity-aware contextual relationships between features significantly enhances the model’s ability to capture complex patterns in large tabular datasets. The architecture design, which includes proximity-aware contextual embeddings and transformer layers, effectively addresses challenges posed by heterogeneous nature of data, including continuous, categorical, and ordinal features.

\subsectionmark{Implications of Findings}
The improved performance of Proximity-Aware Contextual Transformer has several practical implications for Dream11 platform. Accurate predictions of user spending propensity enable more effective upselling strategies and personalized product recommendations, ultimately enhancing user engagement and platform profitability. Additionally, the ability to process large datasets efficiently means that model can be scaled to accommodate growing user base and increasingly diverse data.

\subsection{Why Proximity-Aware Contextual Transformer Architecture Works Well?}
Proximity-Aware Contextual Transformer has emerged as top-performing model in our comparison, showing superior accuracy and consistency in handling tabular data. This section analyzes reasons behind success of the architecture, drawing on insights from recent research.

\textbf{Handling Non-Rotationally-Invariant Features}:
As noted by \citet{chen2023excelformer}, tabular data features are typically non-rotationally-invariant, meaning their significance does not change when data is rotated. In many datasets, a substantial portion of features may be uninformative or irrelevant. When a model incorporates these useless feature interactions, it can harm network generalization. Our model is effective because it mitigates impact of these non-rotationally-invariant and uninformative features.

\textbf{Robustness to Uninformative Features}:
\citet{grinsztajn2022tree} highlighted that neural networks, including traditional MLPs and some transformer models, often struggle with uninformative features, which can lead to inconsistent performance across different datasets. Their research shows that removing uninformative features narrows performance gap between MLPs and other models like FT Transformers and tree-based models. Conversely, adding uninformative features widens this gap, indicating that MLPs are less robust to such features.

Our model, however, demonstrates robustness against uninformative features. This robustness likely stems from its architecture, which efficiently filters out irrelevant information, thus maintaining high performance even in presence of uninformative features.

\textbf{Addressing High Sample Complexity}:
\citet{ng2004feature}, neural networks typically incur a high worst-case sample complexity when dealing with data that contains many irrelevant features. This complexity arises because a rotationally invariant learning procedure must first identify original orientation of features before it can effectively select least informative ones. Proximity-Aware Contextual Transformer architecture addresses this challenge by incorporating mechanisms that reduce influence of irrelevant features early in processing pipeline, thereby lowering sample complexity and enhancing generalization.

\textbf{Embedding Layers and Breaking Rotation Invariance}:
\citet{grinsztajn2022tree} and subsequent studies by \citet{somepalli2021saint} and \citet{gorishniy2023revisiting} emphasize importance of embedding layers in improving performance of MLP and Transformer models. These embeddings, even for numerical features, help by breaking rotational invariance, which is a crucial step for enhancing model performance. Proximity-Aware Contextual Transformer likely benefits from a similar mechanism, where embedding layers or equivalent components break rotation invariance, allowing model to better handle inherent structure of tabular data.

The success of Proximity-Aware Contextual Transformer can be attributed to its ability to effectively manage non-rotationally-invariant features, robustness against uninformative features, and reduced sample complexity. By incorporating architectural elements that break rotational invariance and efficiently filter out irrelevant information, the model achieves superior accuracy and consistency in predictive performance. This analysis underscores importance of designing models that are tailored to unique characteristics of tabular data, ensuring robust and reliable outcomes across diverse datasets.

\subsection{Comparison with Related Work}
The results corroborate findings from previous studies that deep learning models can outperform GBDT models on large-scale tabular data. However, this research advances the field by specifically addressing the proximity-aware contextual relationships between features, a dimension often overlooked in prior work. The success of Proximity-Aware Contextual Transformer aligns with recent trends in leveraging transformer-based architectures for various data types, extending their applicability to tabular data.




%% file: text/conclusion.tex
In this work, we have investigated the performance of deep learning models for tabular data. We have proposed a novel
architecture for tabular data modeling that captures the proximity-aware contextual relationships between input features. We have
evaluated the proposed architecture on a variety of in-house tabular datasets and compared the performance with the
existing deep learning models for tabular data. Our results show that the proposed architecture outperforms the
existing deep learning models for tabular data. The code and all the details of the study are open sourced, we hope
that this work will inspire further research in the field of deep learning for tabular data. We have deliberately
avoided the debate on whether deep learning is universally superior to tree-based models, instead focusing on
demonstrating the specific advantages of deep learning for large-scale tabular data. Future research could explore
hybrid models that combine the strengths of both approaches, particularly in domains with highly heterogeneous data.
Additionally, implementing the Proximity-Aware Contextual Transformer in real-world applications could provide valuable insights into
its practical benefits and potential areas for further optimization.
